\documentclass{article}
\usepackage{graphicx}
\usepackage{placeins}
\usepackage{float}

\usepackage{soul}
\usepackage{xcolor}
\usepackage{arxiv}
\usepackage{caption}

\usepackage[utf8]{inputenc} 
\usepackage[T1]{fontenc}    
\usepackage{hyperref}       
\usepackage{url}            
\usepackage{booktabs}       
\usepackage{amsfonts}       
\usepackage{nicefrac}       
\usepackage{microtype}      
\usepackage{lipsum}
\usepackage{graphicx}
\usepackage{amsmath}
\usepackage{booktabs}
\usepackage{subcaption}
\usepackage{comment}
\usepackage{xcolor}
\usepackage{soul}         
\graphicspath{ {./images/} }

\title{Forecasting MBTA Transit Dynamics: A Performance Benchmarking of Statistical and Machine Learning Models}

\author{
  Sai Siddharth Nalamalpu \\
  Sunset High School\\
  Portland, OR 97229 \\
  \texttt{sainalamalpu123@gmail.com} \\
  \texttt{ORCID ID: 0009-0003-2746-4569} \\
   \And
  Kaining Yuan \\
  Woodbridge High School\\
  Irvine, CA 92604 \\
  \texttt{kevinyuan836@gmail.com} \\
  \texttt{ORCID ID: 0009-0007-4633-8794} \\
  \And
  Aiden Zhou \\
  Western Academy of Beijing\\
  Bei Jing Shi, China 100102 \\
  \texttt{aidenrzhou@gmail.com} \\
  \texttt{ORCID ID: 0009-0008-0989-1651 } \\
  \And
  Eugene Pinsky 
  (corresponding author)\\
  Computer Science Department, Metropolitan College\\
  Boston University, Boston, MA 02215 \\
  \texttt{epinsky@bu.edu} \\
  \texttt {ORCID ID:0000-0002-3836-1851} \\
\
}

\begin{document}
\maketitle



\begin{abstract}
The Massachusetts Bay Transportation Authority (MBTA) is the main public transit provider in Boston, operating multiple means of transport, including trains, subways, and buses. However, the system often faces delays and fluctuations in ridership volume, which negatively affect efficiency and passenger satisfaction. To further understand this phenomenon, this paper compares the performance of existing and unique methods to determine the best approach in predicting gated station entries in the subway system (a proxy for subway usage) and the number of delays in the overall MBTA system. To do so, this research considers factors that tend to affect public transportation, such as day of week, season, pressure, wind speed, average temperature, and precipitation. This paper evaluates the performance of 10 statistical and machine learning models on predicting next-day subway usage. On predicting delay count, the number of models is extended to 11 per day by introducing a self-exciting point process model, representing a unique application of a point-process framework for MBTA delay modeling. This research involves experimenting with the selective inclusion of features to determine feature importance, testing model accuracy via Root Mean Squared Error (RMSE). Remarkably, it is found that providing either day of week or season data has a more substantial benefit to predictive accuracy compared to weather data; in fact, providing weather data generally worsens performance, suggesting a tendency of models to overfit.
\end{abstract}


\section{Introduction}

Public transit forecasting has become increasingly important as cities grow denser and more transportation facilities are developed. Thus, understanding and predicting fluctuations in transit demand is vital for efficient operation, sustainability, and resource allocation. Public transportation is heavily influenced by many factors, in which weather conditions play a crucial role. Understanding how weather impacts the transit system can lead to better planning and ridership satisfaction in the future. Many studies have explored this field and found results on how weather factors such as temperature, precipitation, wind speed, and other conditions affect ridership patterns \cite{1},\cite{2},\cite{3},\cite{4}. For instance, Singhal et al., Guo et al., and Wei et al. studied the relationship between urban transit ridership and weather variations in New York City, Chicago, and Brisbane, through which all of the studies showed some sort of impact of weather on ridership; generally, bad weather tends to reduce ridership. Studying the impact of these same factors in Boston would prove rewarding as Boston's public transit system functions similarly to Chicago and New York City.  

The Massachusetts Bay Transportation Authority (MBTA), serving the Boston area, is similarly vulnerable to fluctuating weather conditions. Over 800,000 people use the MBTA system per month, making it the third-largest public transit system in the United States \cite{5},\cite{6}. Furthermore, Boston is the seventh most densely populated metropolitan area in the United States, making transportation infrastructure a challenging endeavor for city planners \cite{7}. Public transportation is notoriously unreliable in the United States due to many factors that influence its performance.

Currently, these factors create widespread issues with consumer transportation and reliability, leading many to deviate from public transit. MassINC polling group conducted a survey to determine customer satisfaction with MBTA services. They found that with the commuter rail, 46\% of people believe the system is excellent, while in comparison, 35\% rate bus services as at least good, and 28\% felt similarly with the subway and trolley network \cite{7}. When changes of large magnitude occur, the MBTA schedule is largely unreliable, so it is vital that the city of Boston knows how many passengers can be expected on its system, and passengers know when their subways are delayed beforehand. Thus, this paper aims to analyze the impacts caused by weather, season, and day of the week on the subway ridership volume and delays that occur. 

Despite growing interest in public transit prediction, relatively few studies have explored both ridership and delay formation jointly, nor have they examined the relative importance of calendar versus environmental variables in a systematic, multi-model setting. This study investigates how calendar factors (day of week, season) and weather conditions (pressure, average temperature, wind speed, precipitation) influence daily gated-station entries (as a proxy for subway usage) and system-wide delay counts in the MBTA. The MBTA operates subway, commuter rail, and bus services throughout the Boston metropolitan area. 11 statistical and machine learning models are implemented and evaluated, including a novel self-exciting point-process model for predicting delay events. Models are Random Forest Regression, Linear Regression, Ridge Regression, Lasso Regression, Gradient Boost Regression, Support Vector Machine, Multilayer Perceptrons (MLP), K-Nearest-Neighbor Regression, Moving Average (MA), and Poisson Regression. The outcomes from this analysis can help MBTA and other transit authorities in future planning, investments, and the resilience of transportation networks. A public, one-click reproducible GitHub repository for Mac has also been built to facilitate reproducibility (see Appendix).

A narrative literature review was conducted in July 2025, using academic papers that were written in English and published in journals or conferences. The search started by looking for various keywords, including “predicting public transit delays” “predicting passenger ridership” and “train prediction.” Initially, 200+ related papers on this topic were found using ResearchRabbit. Using these papers, the literature review was divided into two parts: subway delays and ridership predictions. 

Past literature focuses on predictions on transit delays between stops (response to \cite{8}), which is effective in letting a passenger know if the train, bus, or subway coming their way is going to be delayed; however, this approach has less real-world practicality. Tracking every bus, train, and subway in a city – or even a state – would take enormous computational power and make the results hard to access. In response to this, our research focuses on aggregate subway delays per day, which requires less computing power and has the benefit of keeping both the local transit authority (in this case, the MBTA) and passengers realistically informed of overall circumstances.

Several studies recognize the myriad of factors that influence delay prediction. Recent work by Zhang et al. \cite{8} extends the delay-prediction field of research beyond the usual single-equation solution. Instead of using a huge model for an entire route, they treat each stop as a separate regression equation and estimate the whole system with a seemingly unrelated regression equations (SURE) estimator, which can catch neglected correlations among residuals, such as driver behavior or traffic shocks. Using six months of bus operation data from Stockholm, Sweden, the stop-specific SURE model achieved RMSEs of 30-42 s and $R^2$ values above 0.92, outperforming an ordinary least squares (OLS) model while still being computationally inexpensive. Zhang et al. highlight that bus operation factors like upstream delay and dwell time are the main cause of delays, whereas calendar and weather factors have weaker effects. 

Srivastava et al. \cite{9} introduce the first rail-specific delay model that combines causal inference with machine-learning uplift modelling. They used a 1.5 million-trip New Jersey transit dataset spanning 2018-2020, and each candidate covariate (such as stop sequence, actual time) was framed as a treatment. The Average Treatment Effect (ATE) and Individual Treatment Effect (ITE) were estimated with a Uplift Tree classifier, which allows them to understand why a forecast changes. They found that stop sequence (ATE roughly equals 0.98) and actual departure/arrival times (ATE roughly equals 0.97) dominated how delays form, more than weather or any other factors. After retraining with only the most influential causal features, the uplift model still attained a 95.6\% accuracy, outperforming XGBoost, Random Forest, and Support Vector Machine (SVM) on the same data. 

Deep learning is also a competitive alternative for ridership forecasting. Demonstrated by Wang et al. \cite{10}, a DST-TransitNet model was proposed, a dynamic spatio-temporal deep learning model that predicts ridership at the station level. DST-TransitNet uses a hybrid deep learning architecture that combines Graph Neural Networks (GNNs), Graph Attention Networks (GATs), and Gated Recurrent Units (GRUs) to dynamically model spatial and temporal features in the data. Unlike previous models that treat stop-to-stop relationships as static, DST-TransitNet calculates dynamic edge weights based on recent ridership, which lets the model understand the changing correlations between stations throughout the day. This is useful when there is fluctuating data due to events like COVID-19 or protests. The model was tested on the Bogota Rapid Transit (BRT) system in Bogota, Colombia, using over five years of 15-minute interval data across 147 stations. DST-TransitNet significantly outperforms state-of-the-art models like LSTM, iTransformer, and DLinear.

Similarly, He et al. \cite{11} developed a Multi-Graph Convolutional-Recurrent Neural Network (MGC-RNN) that incorporates spatial and temporal dependencies and other complex factors. They tackled short-term passenger flow by combining five complementary graphs (network distance, Point-Of-Interest (POI) correlation, network structure correlation, operational information correlation, and recent flow correlation) with a sequence to sequence (seq2seq) LSTM encoder-decoder. The model handles both static and time varying inter-station correlations. Applied to 118 stations on the Shenzhen Metro in China, they showed that holiday, day of the week, and weather variables did not significantly affect the outputs, whereas graph features and recent activity dominate the outputs. 

Tiong et al., consistent with other authors, claims that factors such as crew schedule, rolling stock circulation, infrastructure data, passenger data, weather, peak hours, calendar features, roadwork, and maintenance activities can affect train delays \cite{12}. There are two main areas of improvement that can be taken away from this. Firstly, collection of this sheer amount of data is resource heavy, making it both costly and timetaking. As a result, our project focuses on limiting the number of inputs to make it more real-world friendly. Secondly, existing research overlooks day of the week as a major factor in delay prediction, which our research finds a strong correlation between. Most existing literature looks at more “common sense” factors by asking transportation experts \cite{13} \cite{14}. This would help organizations like the MBTA do their work with minimal cost as the day of the week is easily accessible. 

In regards to passenger ridership per day within the MBTA system, current literature has explored a variety of methods to predict passenger ridership. There is academic literature that focuses on using an ARIMA (Autoregressive Integrated Moving Average), SVM (Support Vector Machine), probabilistic approach (using statistical models), and Deep Learning/Machine Learning approach \cite{15}. Past literature focuses on Deep Learning and ML models; these, on average, perform better than traditional time series models (which focus on a dependent variable changing based on time) \cite{15}. More recent literature focuses on more innovative approaches to this prediction problem, such as using attention based models \cite{16}. Other papers use image analysis of short-term metro traffic as a unique approach to attempt short-term forecasting. Similar to delay prediction, most literature overlooks day of week as a critical variable in predicting passenger flow, which our research presents. In addition, literature does not focus on the models that are the target of our investigation \cite{16}.

The remainder of the paper is organized as follows: Section~\ref{sec:Methodology} presents the methodology of the study, including data analysis and model type discussion. Section~\ref{sec:results_and-discussion} discusses the performance of the various models studied and benchmarks them against each other. Finally, Section~\ref{sec:conclusion} presents a summary of key findings and avenues for further research.

\section{Methodology}\label{sec:Methodology}

\subsection{Statistical Analysis of the data}
The data used for this article came from Meteostat and the Massachusetts Bay Transportation Authority (MBTA).  Meteostat provided the weather data for Boston Logan International Airport, which was used as a proxy for Boston weather during the target time period. The dataset included pressure, wind speed, average temperature, and precipitation.

The MBTA provided all the public transit-related datasets. The "MBTA Service Alerts" dataset \cite{18} provided delay data, while the "MBTA Gated Station Entries" dataset \cite{19} provided ridership data. The CSV file with delay data spans from 2019-01-01 to 2023-07-29. After cleaning and computing daily aggregations for each metric, there are 1671 entries. Though the dataset is classified as "deprecated" by the MBTA, it remains highly accurate and relevant to future regression in 2025 and beyond. 

The gated station entry dataset extends from 2014-01-01 to 2025-06-30, and contains 4199 entries.

After data wrangling, both datasets were processed into 36-column datasets, which contain all data needed for processing (day of week, weather, season, etc.) Processed datasets: \url{https://github.com/icoder178/mbta-prediction-models/tree/main/data}.

\begin{figure}[H] 
    \centering
\includegraphics[width=0.95\textwidth]{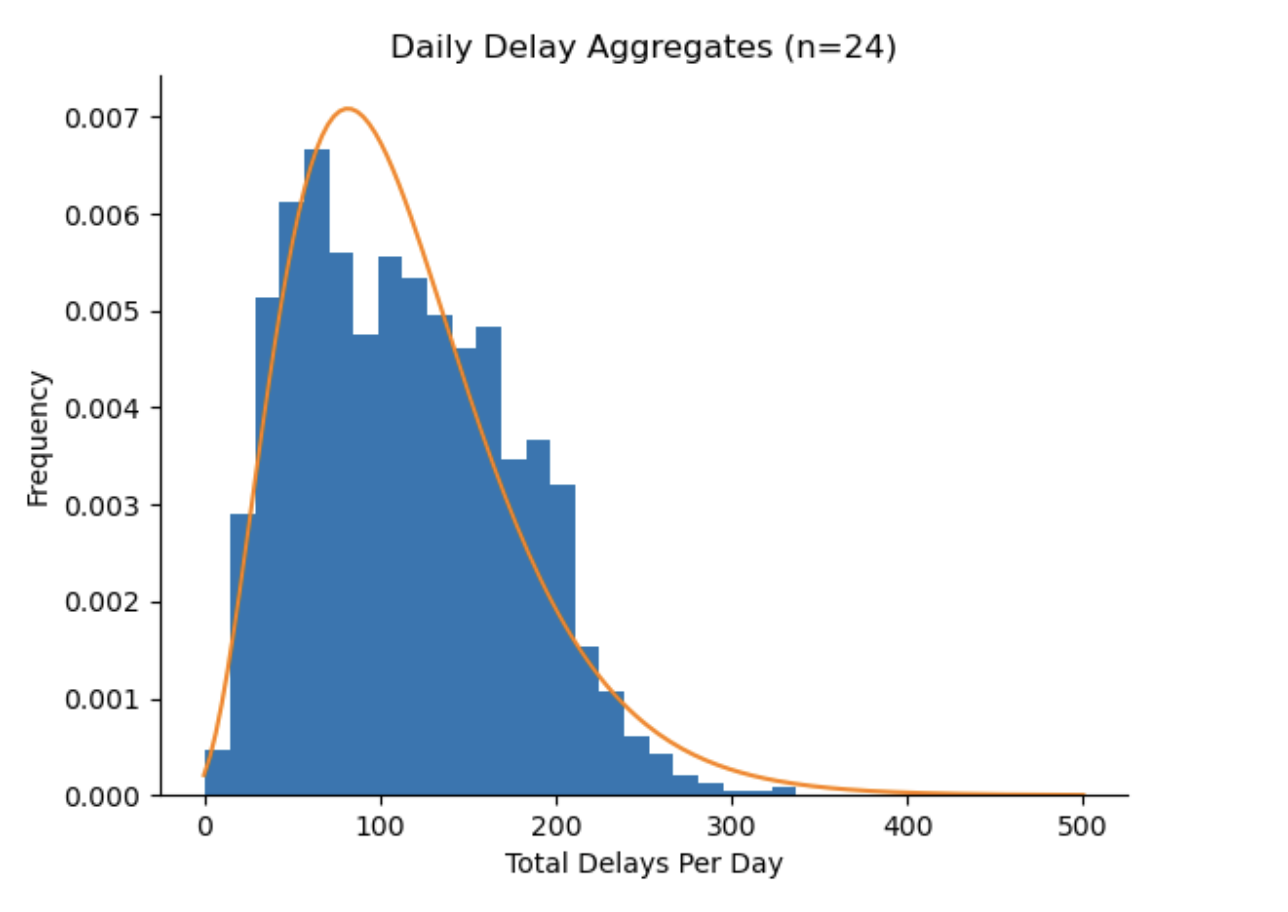} 
    \caption{\textbf{Histogram of Daily Delay Aggregates.} The histogram was created using a bin size of 24, as calculated by Rice’s Rule. The distribution of total delays per day approximates a gamma distribution, which makes it right-skewed. The maximum number of delays on a given day is 337, while the minimum is 1 delay. Figure 1 plots the empirical distribution of total daily delay events (n = 24) and the maximum-likelihood Gamma density that best fits these data. The histogram shows a right-skewed pattern. Most days experience roughly 50 to 170 delays, while there are a relatively smaller number of days with over 250 delays. The fitted three-parameter Gamma distribution (shape = 3.92, scale = 32.03, location = -11.6) highlights that days with extreme aggregate delay counts, though infrequent, are still plausible. The close alignment between the curve and the histogram supports the fact that a gamma distribution is appropriate. On any given day, it is expected that there would be few delays as delays generally occur due to rare instances such as mechanical breakdown, weather, and construction. These rare instances do occur, however, as highlighted by the tail of the graph.}
    \label{fig:delay_aggregates}
\end{figure}
\begin{figure}[H] 
    \centering
    \includegraphics[width=0.95\textwidth]{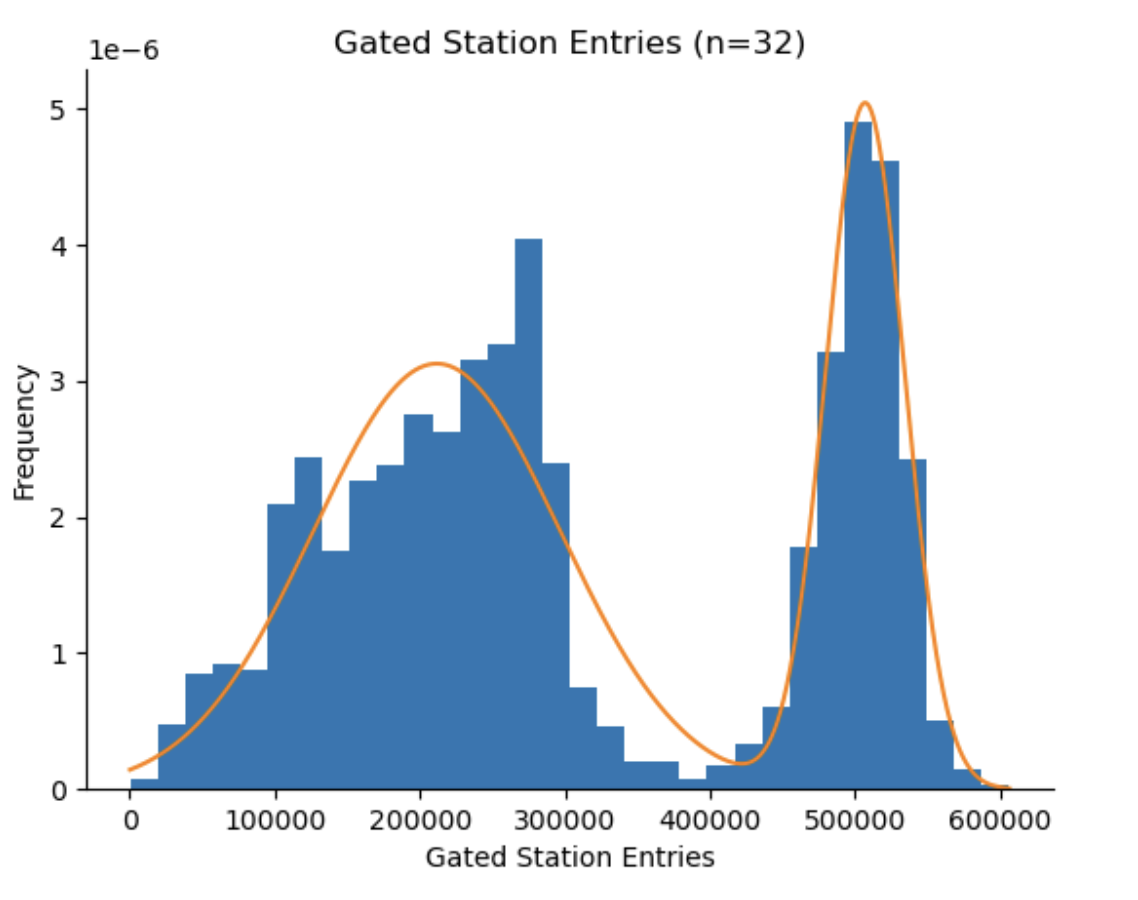} 
    \caption{\textbf{Histogram of Gated Station Entries}
    The histogram was created using a bin size of 32, which was calculated using Rice’s Rule. The distribution of Gated Station Entries follows a bimodal distribution, with two normally distributed peaks. The maximum number of gated station entries on a given day is 606,176, while the minimum was 342. 
    The best fit curve over the histogram visually represents that, during weekdays, more people use the MBTA (e.g., to go to work/school), as represented by the first peak at ~200,000–300,000 entries. The other peak at ~500,000 gated station entries could be due to increased travel on holidays such as Thanksgiving, Christmas, and Independence Day.} 
    \label{fig:gse}
\end{figure}

\subsection{ML Model Comparison}
Our comparative study evaluates how well various machine-learning 
models~\cite{Bishop, Hastle} predict two next-day outcomes for the MBTA system: (i) the total number of recorded service delays (Delay task) and (ii) the total number of gated-station entries (GSE task). For every calendar day, a five-day sliding window is constructed, containing the target series itself plus optional covariates. These optional inputs come from three conceptually distinct groups: day-of-week indicators, seasonal indicators, and four basic weather measurements (pressure, wind speed, precipitation, temperature). After flattening the 5 × k matrix to a single vector, a fixed-length feature tensor suitable for any scikit-learn regressor is created.

To isolate the incremental value of each covariate class, every model architecture is trained on all eight possible combinations of the three groups (with or without day-of-week, with or without season, with or without weather). Because models respond differently to preprocessing, each covariate blend is presented in up to four data representations: raw numeric form, scaled (scikit-learn's StandardScaler() function), one-hot-encoded categories, and a scaled + one-hot version. For any given blend, the representation that yields the lowest test RMSE is maintained, ensuring that an ill-suited encoding does not hinder a model.

Model evaluation follows a bootstrap protocol \cite{21}. One hundred cycles are run, in which the daily records are resampled with replacement; each bootstrap sample has exactly the same length as the original series. In every cycle the first 80\% of chronological dates form the training set and the remaining 20\% serve as the test set (1333 / 333 rows for Delay, 3355 / 839 rows for GSE). Ten architectures are examined: Random Forest, Gradient Boosting, Linear, Ridge, Lasso, Poisson, Support-Vector Regression, multilayer perceptron, k-nearest neighbor, and a simple moving-average baseline, yielding twenty-seven feature-blend experiments per model per cycle. Two headline statistics are recorded for each architecture: “No-Additional-Data” RMSE, obtained when the model sees only the lagged target metric, and “Any-Data” RMSE, the lowest error achieved across all blends and representations. The distribution of these two statistics over the hundred bootstrap replicates provides 95\% confidence intervals, allowing us to judge both the raw skill of each model and the degree to which extra covariates improve its forecasts. Finally, feature importance of the best-performing model (RandomForest in both cases) was then computed through Python’s SHAP library \cite{22}. Violin plots of the most significant 10 features and SHAP values are in Results.

This grid of covariate permutations, data encodings, and bootstrap resampling makes the comparison robust to sampling variability and preprocessing choices, while highlighting which modelling techniques and input signals most effectively predict next-day MBTA ridership and delay totals.

\subsection{Point Process Models}

Temporal Point Process Models are used to depict discrete events that occur over time \cite{23}. Time is a continuous variable. In this case, delayed events over time are modeled, and a Hawkes Process run over those.

A Hawkes Process focuses on the “history-dependent” aspect of a point process \cite{24}. In other words, events in the past can affect the chance of another type of event happening again. This is especially useful to the current problem; for example, when the subway gets delayed, it can lead to additional delays as tracks must handle more traffic than usual. In the present work, a self-exciting Hawkes Point Process Model is implemented.



Each delay is treated as an event in continuous time and summarize the past by the history up to time \(t\), \(\mathcal{H}_t=\{T_i:\, T_i<t\}\).
The (conditional) intensity 
\begin{equation}
\lambda(t\mid \mathcal{H}_t)
= \lim_{\Delta t\downarrow 0}\,
\frac{\Pr\{\text{an event occurs in }[t,t+\Delta t)\mid \mathcal{H}_t\}}{\Delta t}
\label{eq:delay}
\end{equation} 
is the instantaneous rate at which a new delay is expected to occur, given what has already happened \cite{24}. To capture the tendency of delays to arrive in a big amount, a self–exciting Hawkes process is used in which each event temporarily elevates the near–future hazard of subsequent events. With an exponential triggering kernel, the above equation~\eqref{eq:delay} can be re-written as:
\begin{equation}
\lambda(t\mid \mathcal{H}_t)
= \mu + \sum_{T_i<t}\alpha\,e^{-\beta (t-T_i)}\,\mathbb{1}\{t>T_i\},
\label{eq:delaytwo} \end{equation} where \(\mu>0\) is the baseline or background rate,
\(\alpha>0\) is the immediate jump after an event, and
\(\beta>0\) controls how quickly that extra risk decays \cite{25}. Two helpful summaries of equation~\eqref{eq:delaytwo} are the branching ratio \(n=\alpha/\beta\), which is the expected number of directly triggered events by a single event and the half-life \(t_{1/2}=\ln 2/\beta\), which is the time for the post-event boost to halve.

Parameters \(\mu,\alpha,\beta\) are estimated by maximum likelihood from the event times \(\{T_i\}\), using \begin{equation} \ell \;=\; \sum_i \log \lambda(T_i) \;-\; \int_0^T \lambda(s)\,ds .
\label{eq:delaythree} \end{equation} For the exponential kernel, both the compensator \(\int\lambda\) and the event-time intensities can be computed in \(O(N)\) via a simple recursion, so fitting scales well to long histories. In practice, history is truncated beyond a few half-lives when evaluating \(\lambda(t)\), which greatly increases
computation efficiency with negligible loss of accuracy. In addition, approximating parameters through equation~\eqref{eq:delaythree} is necessary to evaluate the performance of the model.

Two targets are evaluated. For next–event timing, the waiting time from “now” to the next delay is forecasted by Monte Carlo simulation \cite{21} from the fitted intensity (Ogata thinning) and score errors in hours. For daily aggregates, the fitted intensity over each calendar day is integrated, $\widehat{y}_d=\int_d^{d+1}\lambda(s)\,ds$, and compared to observed daily totals using RMSE (counts per day). Model performance is checked via cumulative calibration ($N(t)$ versus $\Lambda(t)=\int_0^t \lambda$) and time–rescaling diagnostics.

\section{Results and Discussion}\label{sec:results_and-discussion}
\subsection{Model Performance (excluding Hawkes Model)}
\begin{figure}[H] 
    \centering
    \includegraphics[width=0.95\textwidth]{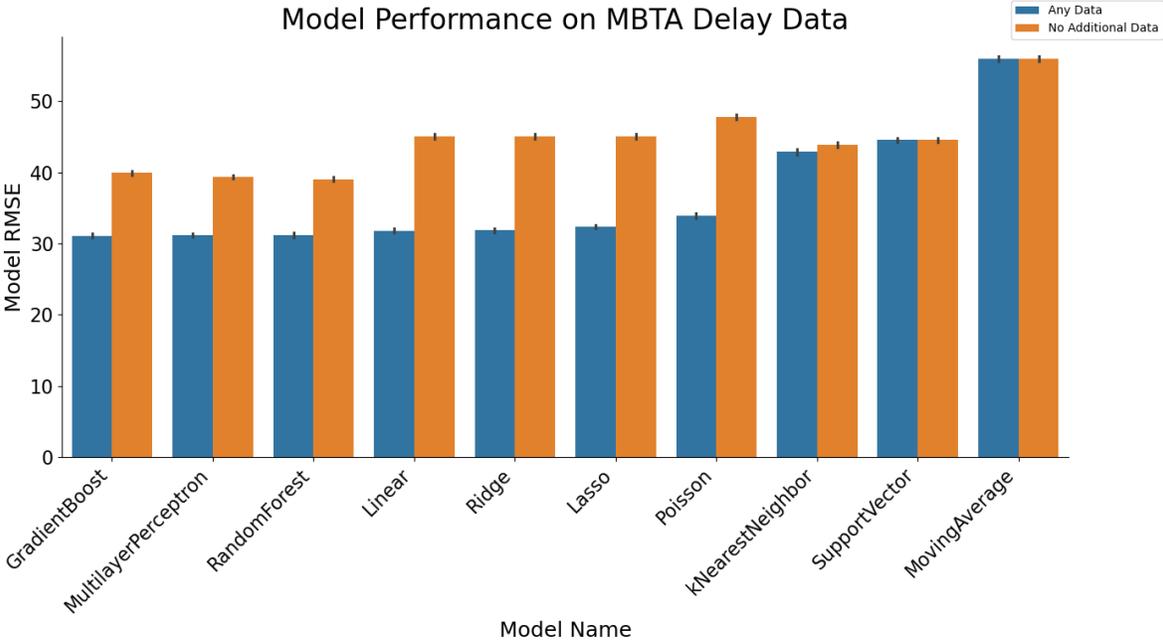} 
    \caption{\textbf{Bar Graph of Model Performance on Delay Data}
    The bar graph shows that 7 out of 10 models predicted significantly more accurately given additional data, including all three highest-performing models. Additionally, the bar graph highlights how random forest regression, multilayer perceptrons, and gradient boost regression appear most suitable for delay prediction, ranking the highest of all models.}
    \label{fig:delay_data}
\end{figure}

\begin{figure}[H] 
    \centering
    \includegraphics[width=0.95\textwidth]{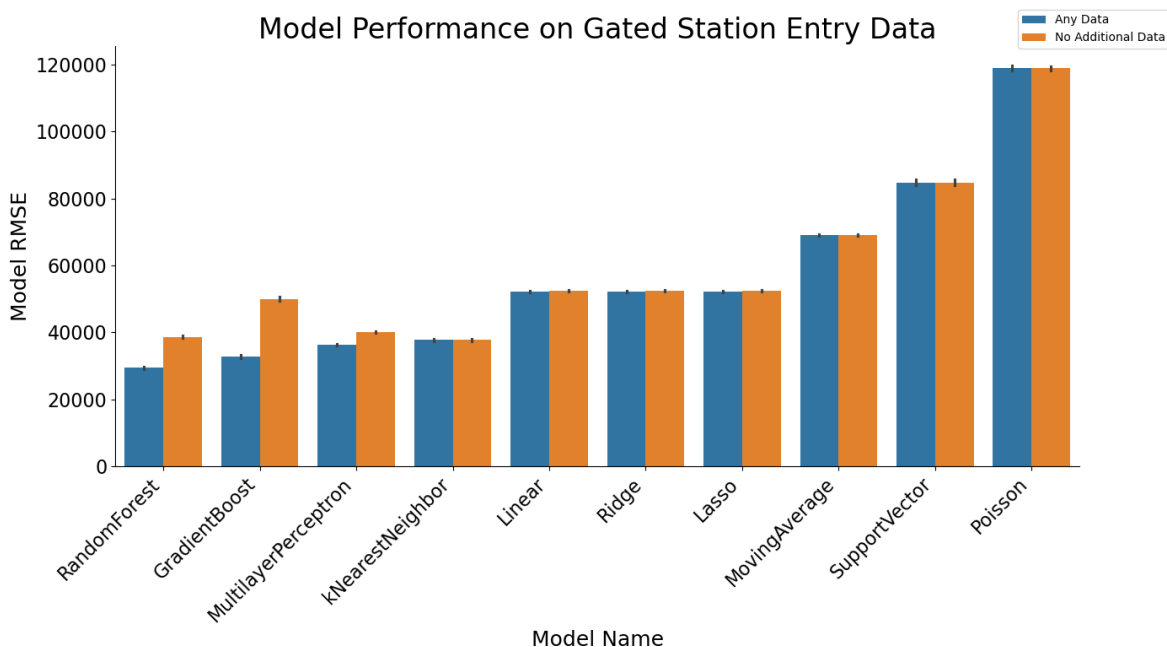} 
    \caption{\textbf{Bar Graph of Model Performance on GSE}
    The bar graph shows 3 of 10 models predicted significantly more accurately given additional data – all three highest-performing models, which are random forest regression, multilayer perceptrons, and gradient boost regression.}
    \label{fig:gse_data}
\end{figure}
Noticeably, both Figure~\ref{fig:delay_data} and Figure~\ref{fig:gse_data} highlight that the same 3 models perform the best for delay prediction and gated station entry prediction, indicating these models tend to perform well for certain time series related tasks.

Additionally, performance for Any Data and No Additional Data in Figure~\ref{fig:delay_data} and Figure~\ref{fig:gse_data} is similar for high-performance models in predicting MBTA delay counts. This suggests there is some factor that these models, regardless of architecture, cannot capture – most likely, some variable that input data does not include; it may also be random noise.
\begin{figure}[h] 
    \centering
    \includegraphics[width=0.95\textwidth]{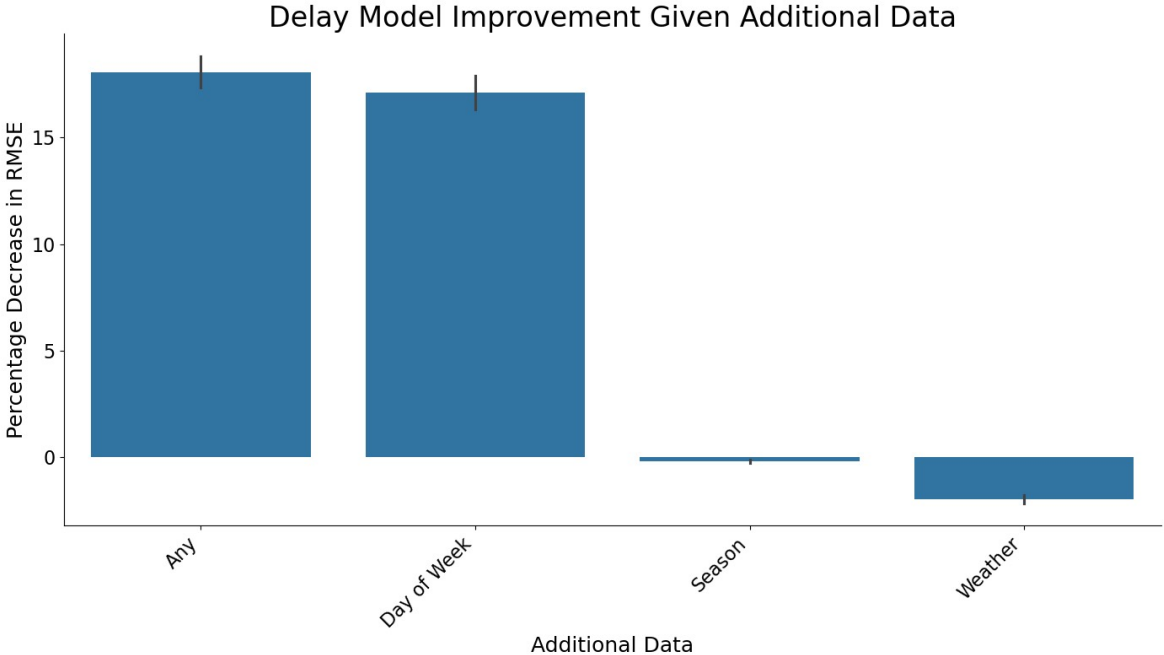} 
    \caption{\textbf{Bar Graph of Delay Model Improvement Given Additional Data}
    Out of all types of data, the figure shows day of week data leads to the greatest decrease in error within delay prediction models while weather data creates the greatest increase in error.
    }
    \label{fig:delay_improvement}
\end{figure}
\begin{figure}[H] 
    \centering
    \includegraphics[width=0.95\textwidth]{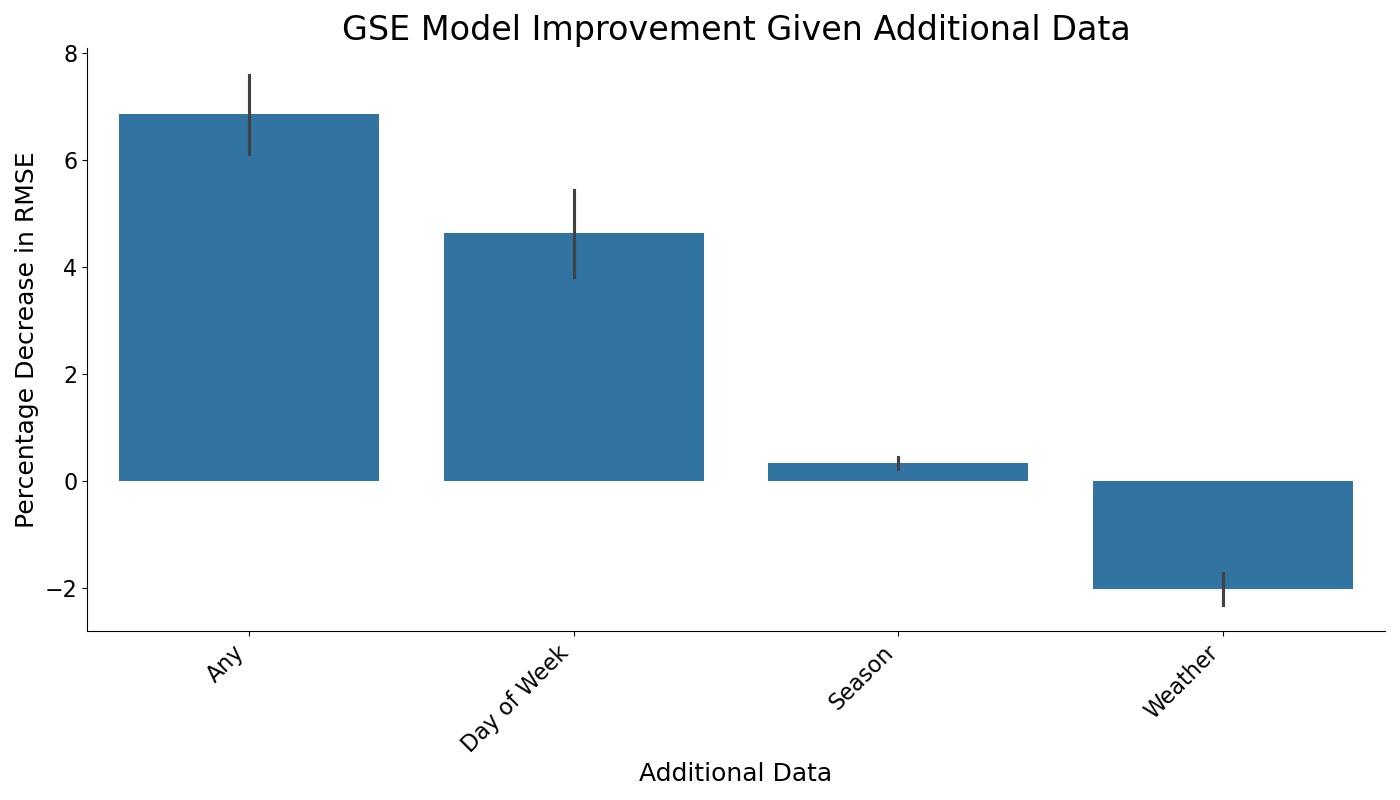} 
    \caption{\textbf{Bar Graph of GSE Model Improvement Given Additional Data}
    The figure shows out of the different types of data, day of week leads to the greatest decrease in error within gated station entry models while weather creates the greatest increase in error.
    }
    \label{fig:gse_improvement}
\end{figure}

As seen in Figure~\ref{fig:delay_improvement} and ~\ref{fig:gse_improvement}, providing additional data (day of week, season, weather metrics) leads to tangible improvements in predictive accuracy, especially for high-performance models. This demonstrates some relationship between these metrics and both MBTA delays and train usage; additional data provides useful information.
\begin{figure}[h] 
    \centering
    \includegraphics[width=0.95\textwidth]{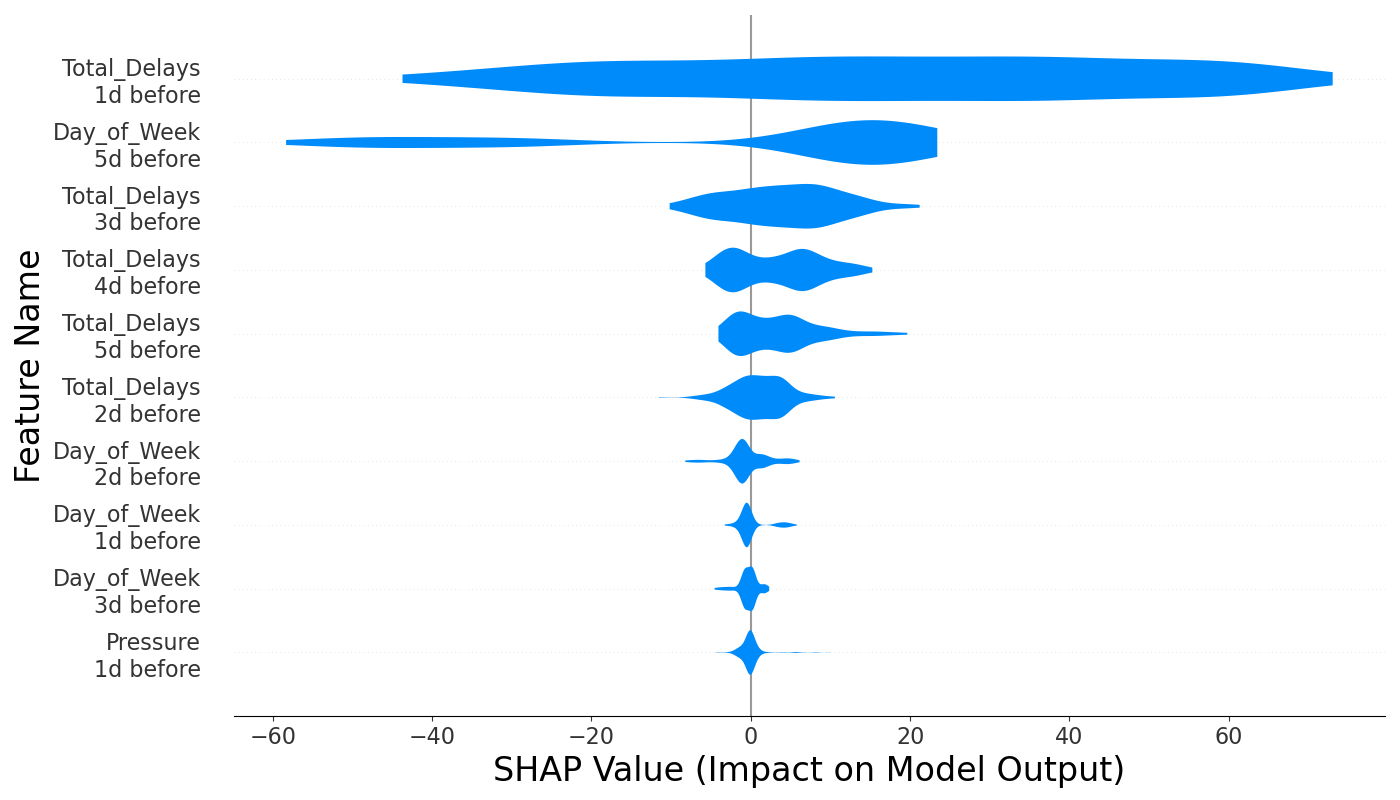} 
    \caption{\textbf{SHAP Feature Importance Values For Delay Prediction}
        The graph highlights the most important factor for delay prediction is total delays 1 day before, while the least important factor is pressure 1 day before.
    }
    \label{fig:delay_feature_importance}
\end{figure}
\begin{figure}[H] 
    \centering
    \includegraphics[width=0.95\textwidth]{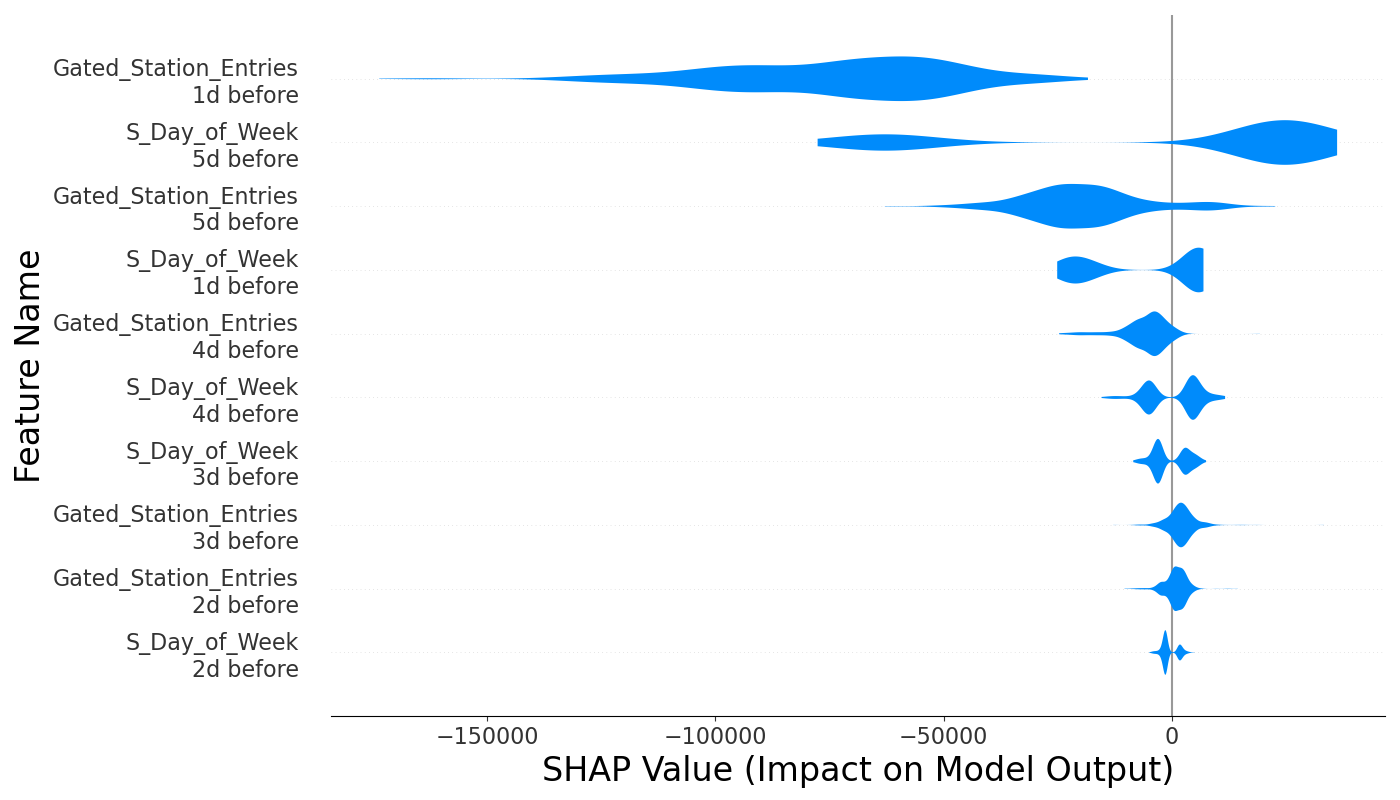} 
    \caption{\textbf{SHAP Feature Importance Values For Gated Station Entry Prediction}
        The graph highlights the most important factor for gated station entry prediction is gated station entries 1 day before, while the least important factor is day of week 2 days before.
    }
    \label{fig:gse_feature_importance}
\end{figure}
Figure~\ref{fig:delay_improvement} and ~\ref{fig:gse_improvement} indicate that including day of week data leads to the greatest improvement in model performance. This is mirrored by the findings in Figure~\ref{fig:delay_feature_importance} and ~\ref{fig:gse_feature_importance}, which show that day of week data is the most important non-target metric factor in model predictions. Thus, this suggests that both delay counts and gated station entries exhibit significant repeating weekly trends that models can leverage to improve performance.

\subsection{Hawkes Point Process Model Performance}

\begin{table}[h]
  \centering
  \caption{Hawkes fit, branching ratio, half-life, and predictive errors}
  \label{tab:hawkes_params}
  \begin{tabular}{lcccccc}
    \toprule
    \(\hat{\mu}\) & \(\hat{\alpha}\) & \(\hat{\beta}\) & \(n=\hat{\alpha}/\hat{\beta}\) &
    \(\tau_{1/2}=\ln 2/\hat{\beta}\) & Daily RMSE & Next-event RMSE \\
    \midrule
    0.438 & 1.884 & 2.087 & 0.90 & 0.33 h & 137.43 counts & 0.670 hour \\
    \bottomrule
  \end{tabular}
\end{table}

\begin{figure*}[h]
  \centering
  \begin{subfigure}[b]{0.49\textwidth}
    \includegraphics[width=\linewidth]{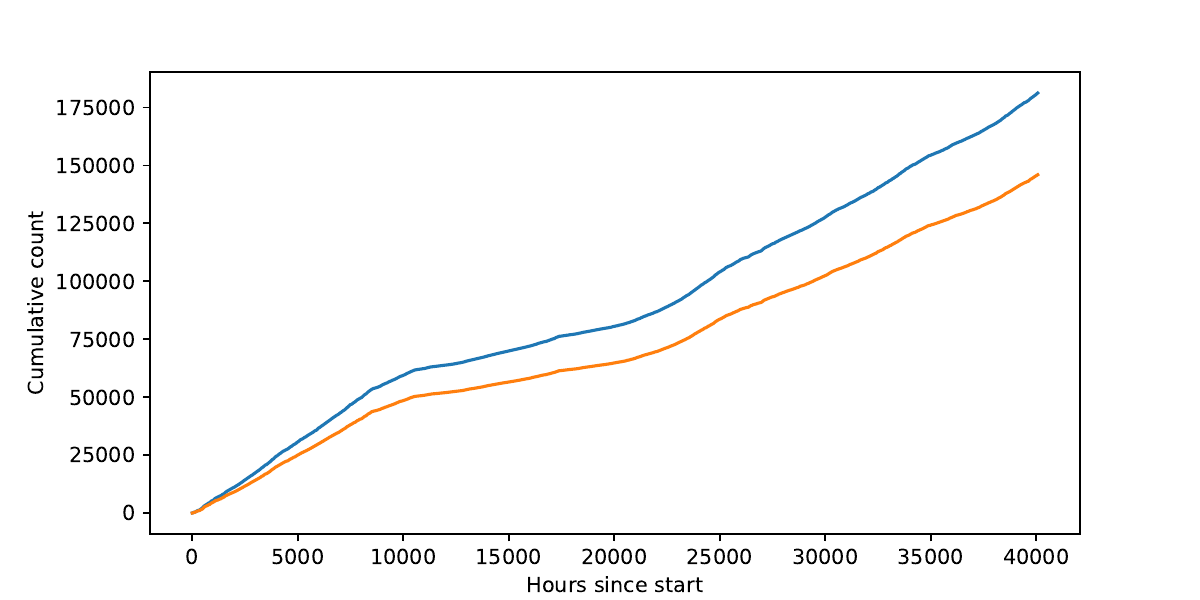}
    \caption{Observed \(N(t)\) vs. expected \(\Lambda(t)\)}
  \end{subfigure}
  \hfill
  \begin{subfigure}[b]{0.49\textwidth}
    \includegraphics[width=\linewidth]{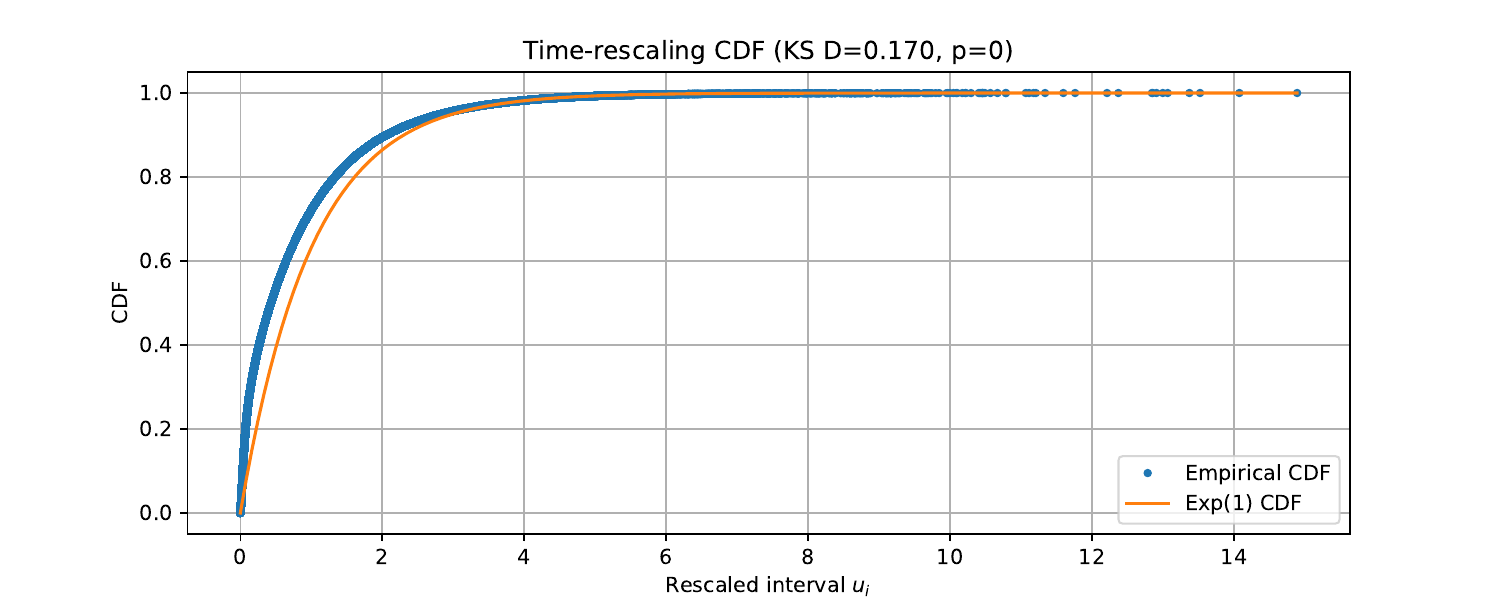}
    \caption{Time-rescaling CDF (ECDF vs. \(\mathrm{Exp}(1)\))}
  \end{subfigure}

  \vspace{0.75em}

  \begin{subfigure}[b]{0.50\textwidth}
    \includegraphics[width=\linewidth]{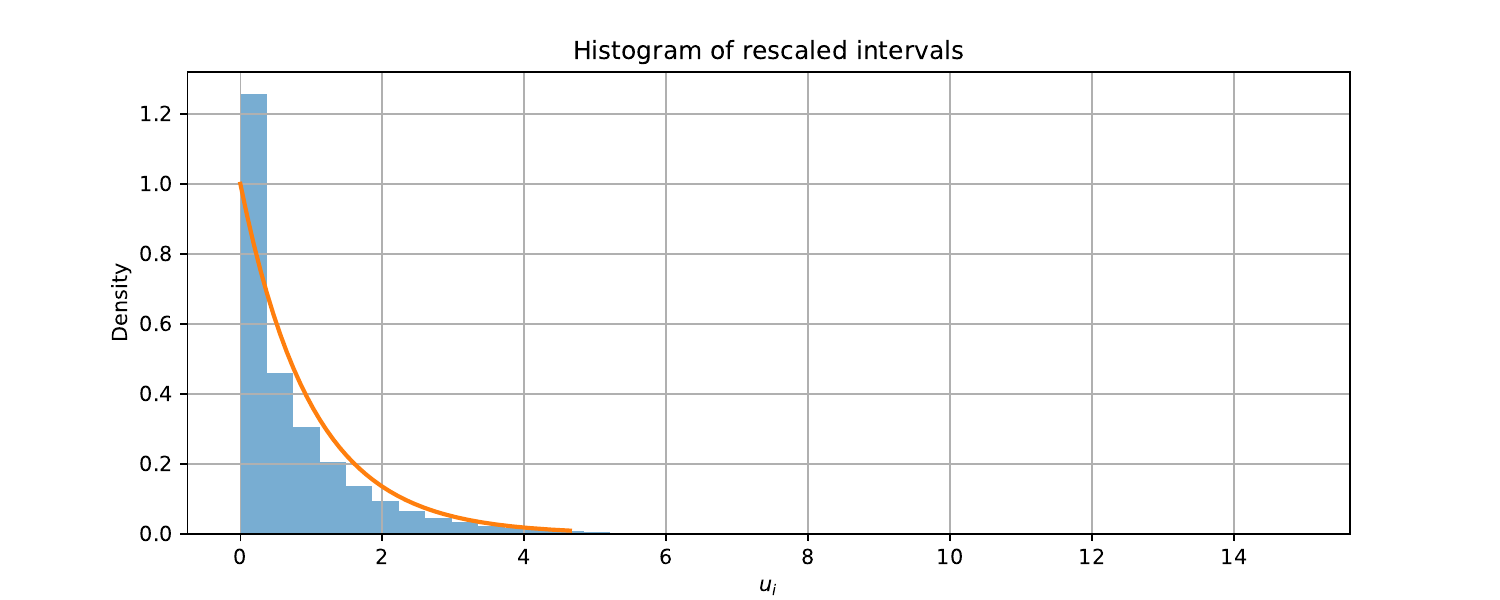}
    \caption{Histogram of rescaled intervals \(u_i\) and \(\mathrm{Exp}(1)\) pdf}
  \end{subfigure}
  \hfill
  \begin{subfigure}[b]{0.49\textwidth}
    \includegraphics[width=\linewidth]{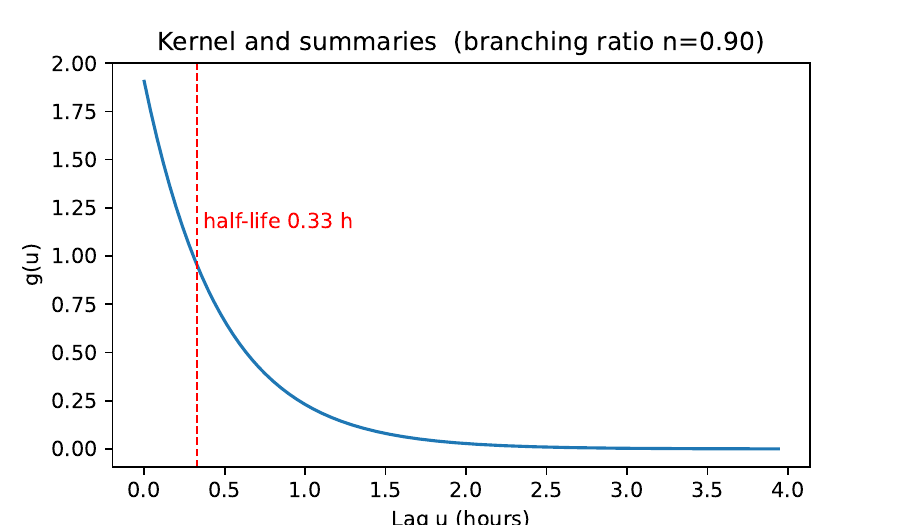}
    \caption{Learned kernel \(g(u)=\hat{\alpha}e^{-\hat{\beta}u}\);
    half-life and branching ratio annotated}
  \end{subfigure}

  \caption{Hawkes diagnostics and learned dynamics. Panels
  A–C assess calibration via cumulative intensity and time-rescaling. Panel D
  summarizes self-excitation strength and decay.}
  \label{fig:hawkes-results}
\end{figure*}

A self–exciting Hawkes process is fit to the sequence of MBTA delay times, using an exponential triggering kernel \(g(u)=\alpha e^{-\beta u}\) and a constant baseline rate \(\mu\). The maximum–likelihood fit yields \(\hat{\mu}=0.438\) events/hour, \(\hat{\alpha}=1.884\), and \(\hat{\beta}=2.087\) (Table~\ref{tab:hawkes_params}).  Interpreting these numbers, the implied branching ratio is \(n=\hat{\alpha}/\hat{\beta}=0.90\), so roughly
90\% of events are explained as knock–ons from previous events rather than from the baseline.  The kernel’s half–life is \(\tau_{1/2}=\ln 2/\hat{\beta}=0.33\) hours,
meaning the extra risk caused by a delay falls by half in about twenty minutes.

Figure~\ref{fig:hawkes-results}a compares the observed cumulative count
\(N(t)\) with the model’s expected cumulative intensity
\(\Lambda(t)=\int_0^t \lambda(s)\,ds\).  The two curves stay close over long
periods, but \(N(t)\) tends to edge above \(\Lambda(t)\) during busy spells, suggesting the model is a bit conservative when bursts cascade.

In Figure~\ref{fig:hawkes-results}b time–rescaling is checked.  If the model is well specified, the rescaled inter–event gaps \(u_i\) should look like i.i.d.\ \(\mathrm{Exp}(1)\).  The empirical CDF rises slightly faster than the \(\mathrm{Exp}(1)\) benchmark, and the Kolmogorov–Smirnov test \cite{27} reports \(D=0.170\) with \(p\approx 0\), so the fit is not perfect.  The histogram in Figure~\ref{fig:hawkes-results}c shows extra mass near zero, consistent with clusters that are tighter than a simple exponential kernel allows.

Figure~\ref{fig:hawkes-results}d plots the learned kernel
\(g(u)=\hat{\alpha}e^{-\hat{\beta}u}\).  With \(n\approx 0.90\) the process is
subcritical (hence stable) yet strongly self–exciting: most delays are
follow–ons rather than baseline events.  The short half–life (about 20 minutes)
indicates that a delay materially raises the chance of another delay within the
next hour, and that this effect fades quickly.

Two prediction tasks are evaluated.  For daily aggregates the fitted intensity over each day is integrated to forecast next–day counts; for next–event delay prediction samples are taken from the conditional intensity (Ogata thinning) to predict the waiting time to the next delay. On our data the Hawkes model achieves an RMSE of \(137.43\) counts/day for the daily task and \(0.670\) hours (40.2 minutes) for the next–event task
(Table~\ref{tab:hawkes_params}). Because units differ, daily RMSE is compared to to count–forecasting baselines and next–event RMSE compared to time–to–event baselines (e.g., empirical inter–arrival summaries). In our study, the Hawkes model trails a tuned moving–average method on daily counts, but it provides calibrated, history–aware next–event predictions and interpretable parameters
that quantify how quickly disruptions trigger follow–up delays.

\section{Conclusion and Future Research}\label{sec:conclusion}
Despite extensive operational planning, the MBTA system still experiences both delays and ridership surges, which reduce passenger satisfaction. Accurate, data-driven forecasting of these patterns is therefore essential not only for efficient resource allocation but also for informing riders about service reliability. Thus, this work benchmarks multiple statistical and machine learning methods for performance comparison. Our analysis demonstrates that simple calendar features such as day-of-week and seasonality provide the strongest signal for both daily ridership and delay counts on the MBTA network. Across ten different modeling approaches, the models that incorporated day-of-week flags and seasonal indicators consistently outperformed versions relying solely on weather variables. In fact, adding weather data by itself slightly degraded performance, confirming that passenger flows and delay patterns are driven more by human schedules than by meteorological factors.

Among the machine-learning techniques benchmarked, ensemble methods and multilayer perceptrons achieved the lowest RMSEs. Random Forest and Gradient Boosting models, in particular, were the most robust and are promising candidates for day-ahead forecasting of MBTA outcomes. Complementing these aggregate predictors, a self-exciting Hawkes point process is fit to the event times of delays. The fitted model revealed strong but short-lived self-excitation (branching ratio $\approx$ 0.9; half-life $\approx$ 0.33 hours), meaning one delay noticeably raises the chance of another within the next hour, then the effect fades quickly. While the Hawkes model underperformed simple moving-average baselines on day-level count prediction, it provided promising next-event prediction with interpretable results, and its diagnostic plots (cumulative calibration and time-rescaling) highlighted when bursts were under- or over-estimated. Practically, this suggests a great distinction: calendar-driven ML models remain best for day-ahead volumes, while the Hawkes model is most useful for real-time dispatch and crowding reduction, where quantifying short-term clustering of disruptions matters.

It is worth noting that simpler time series models (e.g. ARIMA \cite{26}) were not used within this research. This was primarily as poor performance of the Hawkes model (RMSE >130 counts/day, compared to RMSE <60 for a simple moving average) was observed on handling daily aggregates. This was likely because, given the number of events each day, any statistical model regresses to its mean due to the law of large numbers. Other time series models would encounter similar issues, making an investigation into using them for handling daily aggregates not worthwhile.

Looking ahead, several directions remain plausible to extend our work. Our current analysis aggregates most inputs at the daily level; moving to hourly or station-level resolution could uncover localized effects and strengthen both the ML and point-process components. Incorporating operational variables—train assignments, crew shifts, maintenance windows, unplanned incidents, and major events—should explain residual variation and improve forecast accuracy. Finally, embedding spatial network structure (e.g., line topology and transfer hubs) into both the predictive models and the Hawkes kernel could capture how disruptions propagate through the system. With these enhancements, future models can deliver more precise guidance to transit agencies and their riders, combining reliable day-ahead planning with actionable, real-time risk estimates of cascading delays.

In short, conclusions are:
\begin{itemize}
\item Models like Random Forest and Gradient Boost appear most effective at predicting daily aggregates in real circumstances.
\item Fitting a Hawkes point process suggests that chance of delay increases immediately following a prior delay.
\item In the future, research extensions include handling data with increased resolution and incorporating more variables to improve predictive accuracy.
\end{itemize}

\noindent
\section*{Declarations}

\medskip
\noindent
{\bf Conflict of Interest: } We declare that there are no conflicts of interest regarding the publication of this paper.

\medskip
\noindent
{\bf Author Contributions: } All three authors contributed equally.

\medskip
\noindent
{\bf Funding: } This research was conducted without external funding. All study aspects, including design, data collection, analysis, and interpretation, were carried out using the resources available within the authors' institution.

\medskip
\noindent
{\bf Acknowledgements:}
We thank Ursula Imbernon, Patrick Bloniasz, Tharunya Katikireddy,  Tejovan Parker, Zhengyang Shan, and staff at the Research in Science and Engineering (RISE) Summer 2025 Program at Boston University for their support and guidance.

\medskip
\noindent
{\bf Data Availability:}
Most relevant data and analysis are available via:

\medskip
\noindent
\url{https://github.com/icoder178/mbta-prediction-models.git}

\bibliographystyle{unsrt}  
\nocite{*}

\bibliography{references}

\end{document}